\documentclass{article}
\usepackage{ijcai20}

\usepackage{times}
\usepackage{soul}
\usepackage{url}
\usepackage[hidelinks]{hyperref}
\usepackage[utf8]{inputenc}
\usepackage[small]{caption}
\usepackage{graphicx}
\usepackage{amsmath}
\usepackage{amsthm}
\usepackage{booktabs}
\usepackage{algorithm}
\usepackage{algorithmic}
\usepackage{xcolor}

\title{Federated Learning for Resource-Constrained IoT Devices: \\Panoramas and State-of-the-art}

\author{
Ahmed Imteaj$^{1,2}$\and
Urmish Thakker$^{3}$\and
Shiqiang Wang$^{4}$\and
Jian Li$^{5}$\And
M. Hadi Amini$^{1,2}$\footnote{Contact Author: hadi.amini@ieee.org, www.solidlab.network}\\
\affiliations
$^1$School of Computing and Information Sciences, Florida International University\\
$^2$Sustainability, Optimization, and Learning for InterDependent networks laboratory (solid lab)\\
$^3$Arm ML Research\\
$^4$IBM T. J. Watson Research Center\\
$^5$Binghamton University, the State University of New York
\emails
{\small aimte001@fiu.edu,
urmish.thakker@arm.com,
wangshiq@us.ibm.com,
lij@binghamton.edu, moamini@fiu.edu}
}

\newcommand{\citet}[1]{\citeauthor{#1}~[\citeyear{#1}]}

\begin{document}

\maketitle

\begin{abstract}
Nowadays, devices are equipped with advanced sensors with higher processing/computing capabilities. Further, widespread Internet availability enables communication among sensing devices. As a result, vast amounts of data are generated on edge devices to drive Internet-of-Things (IoT), crowdsourcing, and other emerging technologies. The extensive amount of collected data can be pre-processed, scaled, classified, and finally, used for predicting future events with machine learning (ML) methods. In traditional ML approaches, data is sent to and processed in a central server, which encounters communication overhead, processing delay, privacy leakage, and security issues. To overcome these challenges, each client can be trained locally based on its available data and by learning from the global model. This decentralized learning approach is referred to as federated learning (FL). However, in large-scale networks, there may be clients with varying computational resource capabilities. This may lead to implementation and scalability challenges for FL techniques. In this paper, we first introduce some recently implemented real-life applications of FL. We then emphasize on the core challenges of implementing the FL algorithms from the perspective of resource limitations (e.g., memory, bandwidth, and energy budget) of client devices. We finally discuss open issues associated with FL and highlight future directions in the FL area concerning resource-constrained devices.
\if
{\color{black} Nowadays, devices are} equipped with advanced sensors with higher processing/computing capabilities. Further, widespread Internet availability enabled communication among sensing devices.  As a result, {\color{black}vast} amounts of data are generated on edge devices to drive the Internet of Things (IoT), crowd-sourcing, and other emerging technologies. {\color{black}The collected large data can be pre-processed, scaled,  classified, and finally, predicted for any future event using  machine learning (ML) methods.} In traditional  ML approaches, data is sent to and processed in a central server, which encounters communication overhead, processing delay, privacy {\color{black}leakage}, and security issues. {\color{black} To overcome these challenges, each client can be trained locally based on its available data and by learning from the global model. This decentralized learning structure is referred to as Federated Learning (FL).} However, in large-scale networks, there may be clients with varying computational resource capabilities. This may lead to implementation and scalability challenges for FL techniques. {\color{black} In this paper, we first introduce some recently implemented real-life applications of the FL. We then emphasize on the core challenges of implementing the FL algorithms from the perspective of resource limitations (e.g., memory, bandwidth, and energy budget) of client clients. We finally introduce open issues that are associated with the FL and highlight future directions in the FL area concerning resource-constrained devices.\\
\fi
\end{abstract}

\section{Introduction}
The Internet-of-Things (IoT) penetration rate has recently expanded prodigiously due to the integration of billions of connected IoT devices. Such IoT edge devices include different kinds of robots, drones, and smartphones, which have limited computation and storage capabilities and can communicate with remote entities via wide-area network (WAN). The data generated from these devices at the network edge is increasing exponentially. Due to bandwidth and privacy concerns, it is infeasible to send all locally collected data to the server. However, many IoT applications require the prediction and classification of data, for which machine learning (ML) models need to be trained using data collected by multiple devices. The question is: \emph{how to train ML models from the decentralized data at resource-constrained IoT devices?}

To address the above problem, we need to devise an approach through which the learning process can be accomplished without exchanging raw data between client devices. \emph{Federated learning (FL)} is a technique that fulfills this purpose.
With FL, each device can attain a global view and predict a situation that is seen in another device. For example, we can consider a scenario where multiple drones are placed at different locations, and each drone observes vehicles that are passing through their respective ends. If a drone observes a vehicle and gets trained by this, other drones can learn without observing that vehicle through the FL approach. Applications of FL covers a wide spectrum of domains, e.g., word prediction by training models on local edge devices and not sharing sensitive information to the central server \cite{hard2018federated},  adaptive keyword spotting using a locally-trained voice assistant \cite{leroy2019federated}.

Challenges in implementing FL in the presence of heterogeneous hardware in a system are discussed in \cite{lim2019federated,park2019distilling}. Due to various resource constraints (i.e., limited capacity of communication, computation, storage, etc.) at different types of client devices, the clients cannot be treated uniformly, and additional care needs to be taken to handle such heterogeneity. {\color{black}A feasibility study to implement FL on the low-processing unit (e.g., Raspberry Pi)} is given by \citet{das2019privacy}, where they also discussed the potential challenges to detect emotion from sound. Besides, FL on battery-powered devices is studied in \cite{xu2019exploring}, where a two-layered strategy to train the model of battery-powered devices is used. In the first layer, the candidates who have sufficient power to carry out the training process are selected, and in the second layer, FL is applied on those selected battery-powered devices considering local energy optimization. This approach eliminates the problems related to straggler clients and helps improve the overall performance of the training process. 

A blog on training models on edge devices \cite{pete} focuses on approaches for model training and crucial aspects that matter while infusing deep learning models on-device. \citet{bonawitz2019towards} designed a scalable FL system that highlighted significant components of the system, including a high-level view of device scheduling. This study focused on approaches that we can take while training models and the crucial aspects of training FL models at scale. 

{\color{black}A resource-aware FL framework considering heterogeneous edge clients is proposed by \citet{xu2019elfish}. In this work, they eliminate the straggler clients using an optimization technique named `soft-training' that dynamically masks different neurons based on model updates, and their proposed aggregation scheme speeds up the collaborative convergence.
Moreover, a control algorithm that adapts the communication and computation trade-off is presented by \citet{wang2018edge}. They minimize the loss function for a predefined resource budget based on analyzing the impact of different configurations on the convergence rate.

However, there is no comprehensive survey on FL challenges and issues from the perspective of resource-constrained IoT clients. The main contribution of this paper is that we analyze the core challenges, potential solutions, open questions, and future pathways that need to be considered and addressed in the implementation of FL  on resource-constrained IoT devices in the network.

The rest of this paper is organized as follows. Section 2 introduces the background of FL. In Section 3, we discuss some applications of FL. In Section 4, we introduce the core challenges to implement FL particularly for resource-constrained devices, and outline some potential solutions to address these challenges. Further, we highlight the open issues of FL algorithms and hardware developments in Section 5. In Section 6, we list some future directions, which is followed by Section 7 that concludes the paper.

} 

\section{Background of Federated Learning}
The FL algorithm aims to learn a single, global model from local data collected by millions of distributed devices. It learns a model under various resource-constraints of devices, where the model is trained locally and intermediate model updates are shared with the cloud (server) periodically. The overall goal is to minimize a training objective (loss) function which can be written as follows \cite{mcmahan2016communication}:
\begin{equation}
    \min_{z} F(z):= \sum_{i=1}^n P_iF_i(z),
    \label{eq:loss}
\end{equation}
where $n$ is the number of devices, $P_i\geq 0$ 
defines the relative impact of each local device, satisfying $\sum_{i} P_i = 1$. $F_i$ is the objective function of the $i$-{th} local device, 
which can be defined as
$F_i(x)$ = $\frac{1}{s_i}\sum_{j_i=1}^{s_i} f_{ji}(z; x_{ji}, y_{ji})$, where $s_i$ is the number of locally available samples, $P_k ={1}/{s}$ or $P_k ={s_i}/{s}$, and total samples $s = \sum_{i} s_i$.

The FL terminology was first introduced by \citet{mcmahan2016communication} with an algorithm called federated averaging (FedAvg). FedAvg includes multiple rounds of communication between clients and the server which are interleaved by multiple local model update steps at each client. 
The clients do not share raw data due to security and privacy considerations. In this way, FL preserves data privacy since only the statistical summary or model weights are shared with the server.

\begin{figure}[t]
\setlength{\belowcaptionskip}{-20pt}
\begin{center}
  \includegraphics[width=0.8\linewidth]{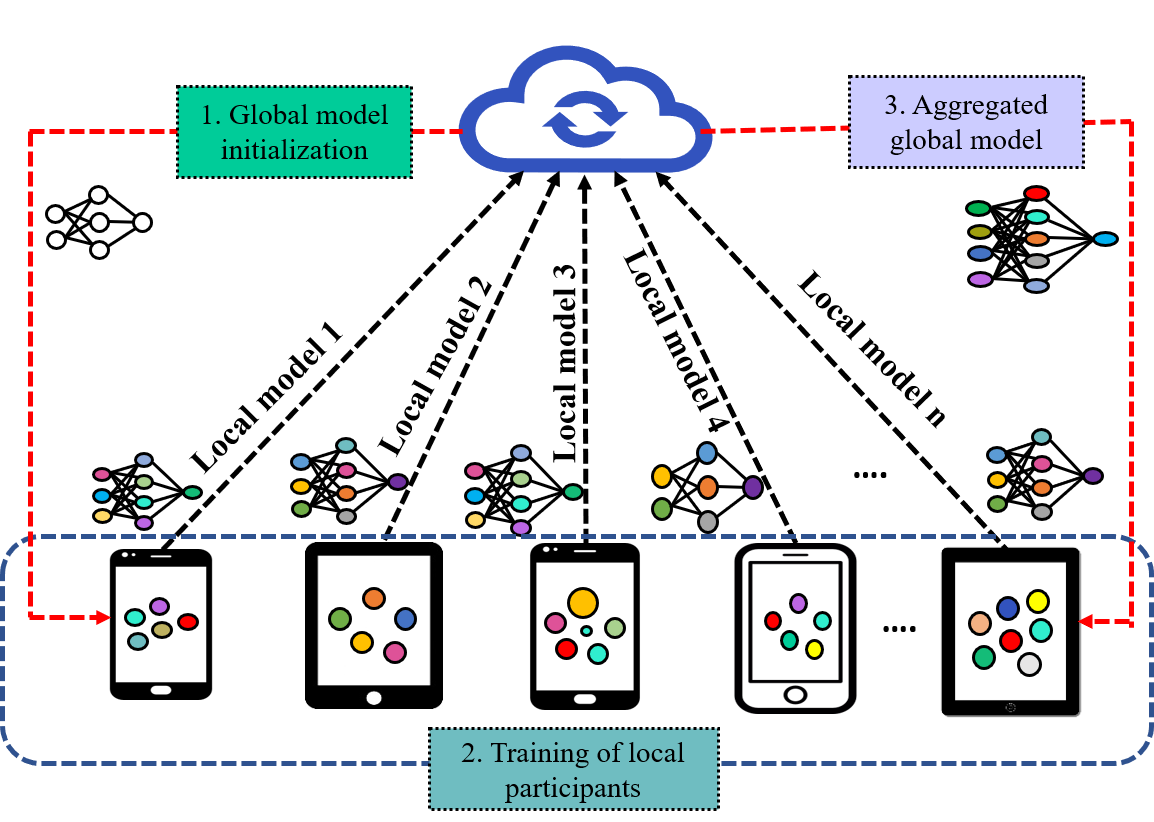}
    \caption{FL procedure considering $N$ participants.}
    \label{fig:l}  
    \end{center}
\end{figure}
FL generally includes three steps as shown in Figure~\ref{fig:l}:

\noindent\emph{\textbf{Step 1 (Initialization of training task and global model)}}: In the initial phase, the central server decides the task requirement and target application. An initial global model is generated by using hyper-parameters and maintaining a training procedure (e.g., learning rate) specified by the server. Then the server broadcasts the initialized global model $W_G^0$ to the selected local participants.

\noindent\emph{\textbf{Step 2 (Local model update)}}: 
Each participating client in the network has a collection of data from where it performs local model update. Upon receiving the global model $W_G^t$, where $t$ denotes the $t$-th iteration, each client $i$ updates its model parameters $W_i^t$ with the goal of  
finding optimal parameters $W_i^t$ that minimizes the local loss function $F_i(W_i^t)$.

\noindent\emph{\textbf{Step 3 (Global aggregation)}}: 
The central server aggregates the local updates received from all clients and generates an aggregated updated global model $W_G^{t+1}$. This latest global model is then sent back to clients who contributed to generating this new model. The goal of the central server is to minimize the global loss function 
$F(W_G^t)$ as defined in (\ref{eq:loss}).

Steps 2 and 3 are repeated until the central server attains target training accuracy or reaches a convergence.

\section{Federated Learning Applications}
FL suits best in applications where data within the device is more significant than data located in the server. Current applications related to FL are mostly based on supervised learning, typically utilizing labels retrieved from user activities (e.g., button click, keyboard type, etc.). In this section, we briefly discuss some applications based on FL.

    \noindent $\bullet$ \textbf{Smart Healthcare:} Smart healthcare involves sensitive data, and it needs to train models on-device. For instance, heart attack situations can be predicted locally for end-users with wearable devices \cite{huang2018loadaboost}. \if{\color{black}For instance, heart-attack risk prediction from wearable devices could be a potential application \cite{huang2018loadaboost}, where the heart-attack situation would be predicted locally for each user.}\fi  \citet{yang2018applied} suggested that if all medical centers cooperate to form a large dataset by sharing their data with proper labeling, the performance of ML model would be remarkably improved. The combination of FL and transfer learning is a preeminent way to achieve this goal.
    
     \noindent $\bullet$ \textbf{Recommendation System:} It is a widely used method that depends on information sharing among users, which may cause privacy leakage. To handle this, \citet{chen2018federated} proposed a federated meta-learning based approach, where local devices share their algorithm instead of data or model with the central server. It eliminates the risk of privacy leakage and leads to proper training of the model on the  local devices.
    
     \noindent $\bullet$ \textbf{
     Next-word Prediction:} An on-device, distributed framework for next-word prediction for smartphones is proposed by \citet{hard2018federated}. They trained the local client devices using the FedAvg algorithm and observed higher prediction recall compared with other approaches that transferred some sensitive information to the server for learning.
    
     \noindent $\bullet$ \textbf{Keyword Spotting:} An embedded speech model, i.e., wake word detector, is proposed in \cite{leroy2019federated}, where an experiment using ‘Hey  Snips’ keyword is conducted based on a crowd-sourced dataset. For keeping user's speech private, they applied the FL strategy. 

     \noindent $\bullet$ \textbf{On-device Ranking:} An on-device ranking of search results is implemented in  \cite{bonawitz2019towards} without conducting expensive calls to the server. This avoids issues related to constrained resources, and sensitive information remains on the device.  The system can label the user interaction with the ranking feature by observing the user’s preferred selected item from the ranked list during the interaction period.

     \noindent $\bullet$ \textbf{Relevant Content Suggestions for On-Device Keyboard:} 
    A content suggestion approach for on-device smartphone keyboard was presented 
    in \cite{yang2019federated}, 
    where value is added to the users by suggesting related contents. For instance, while typing on a keyboard, it can suggest related contents by getting triggered based on the assigned value learned through 
    on-device training.
    

\section{Core Challenges to Implement Federated Learning on Resource-Constrained Devices}
\label{sec:4}
In real-world IoT environments, available clients may have heterogeneous hardware (e.g., memory, computational ability) and variant resource-assets  (e.g., energy budget). As a result, we cannot consider all available clients uniformly as the client's behavior depends on resource-availability. Besides, heterogeneity in hardware can preclude weak clients from participating in FL. 

\begin{figure}[htb!]
\setlength{\belowcaptionskip}{-20pt}
\begin{center}
  \includegraphics[height=0.68\linewidth]{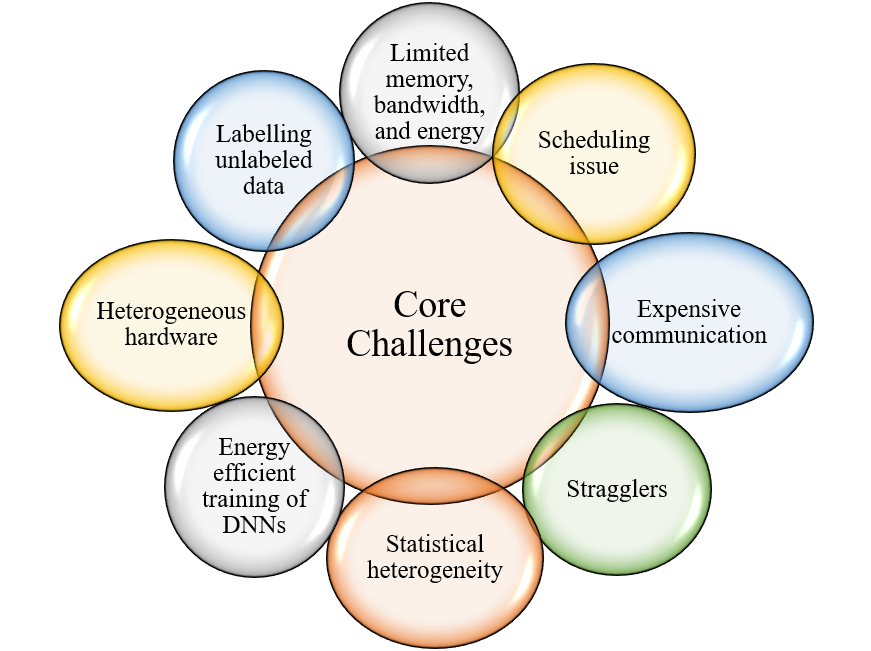}
  \vspace{-0.1in}
    \caption{Core challenges associated with resource-constrained IoT clients in the FL system.}
    \label{fig:2}  
    \end{center}
\end{figure}

Extensive surveys were conducted on the architecture and training process of IoT edge networks and FL \cite{li2018learning,cui2018survey,park2019distilling,yang2019federated}, but challenges of FL systems due to resource limitation were not comprehensively discussed. In this section, we discuss the core challenges (see Figure~\ref{fig:2}) associated with the implementation of FL for resource-constrained IoT devices.

\subsection{Limited Memory and Energy Budget}
The agents participating in FL may have limited memory capacity, constrained computational ability, and bounded energy budget. While reduced computational capabilities imply that it takes more time to process data, limited memory capacity makes the device prone to over-flooding. These situations can lead to more expensive communication (see Section~\ref{subsec:expensiveCommunication} for more details) and curtail the overall performance of the system. \citet{das2019privacy} analyzed hardware limitation challenges in the implementation of FL by considering Raspberry Pi clients. They studied the feasibility of implementing FL on resource-constrained edge devices.  Furthermore, necessary hardware requirements are highlighted in \cite{hard2018federated} during the implementation of next word prediction on keyboard. In terms of hardware requirement, the devices must have at least 2 gigabytes of memory availability, whereas some microcontrollers have very limited memory. Such memory limitation of clients can be managed by storing limited sizes of data and this will help the resource-bounded clients to process those data locally. After a certain period, data within a client can be aggregated and backed-up to avoid unexpected overflow of memory. In this regard, a novel FL-based approach can be adopted through which shards of data are distributed to the clients to obtain the target model quickly \cite{haddadpour2019trading}. According to resource availability, we can also choose proficient clients that have higher bandwidth, better processing ability, greater memory size, and higher energy budget to participate in FL, and clients with resource shortage will not participate.

\subsection{Expensive Communication}
\label{subsec:expensiveCommunication}
The necessity of training local model can be motivated by insufficient communication bandwidth to broadcast local data to a server for central computing. In FL, the server interacts with clients for getting updates and based on local model training, after which the server disseminates an updated global model.  When we consider resource-constrained clients with limited bandwidth and transmission power, it is challenging to utilize those resources prudently for reaching a convergence. Furthermore, FL systems may comprise millions of devices with various degrees of resources, and local computation within the devices can be faster than the network communication \cite{huang2013depth}. Although frequent interaction between the server and the clients can help us to attain a target model swiftly, it is costly to perform communication repeatedly. We need to consider this trade-off while designing optimization algorithms to make proper use of limited resources. \citet{ma2017distributed} discussed the trade-offs between communication expenses and optimal resource utilization, but they did not study the complexity of the local problem's solution. We need to devise a way to achieve the target model by sending a compressed size of message \cite{konevcny2016federated} and by carrying out a minimal number of communication round between the server and local clients.  

\subsection{Heterogeneous Hardware}
In FL, training can run on multiple devices, each coming from different vendors or belonging to a different generation of products. It creates a network of devices with varying computing and memory capabilities and different battery lives. Therefore, training efficiency may vary significantly across client devices, and considering all clients with the same scale does not provide us an optimal solution. \citet{li2019federated} discussed why FL should be aware of heterogeneous hardware configurations. We need to select clients for training purposes based on system requirements. However, due to strict cost and energy requirements, only a few clients might end up meeting the required criterion. It is possible that most of the proficient clients go out of the network, and existing clients do not fulfill system requirements. Hence dealing with resource-constrained heterogeneous devices is a challenge.

\subsection{Energy Efficient Training of DNNs}
\label{subsec:energyEfficientTraining}
Deep neural networks (DNNs) are widely used, especially in implementing artificial intelligence-based applications. It is difficult to perform DNNs on resource-constrained clients as we need to ensure the required processing capability and energy availability. \citet{wu2018training} discussed enabling training on local devices. They demonstrated a way to carry out both inference and training with comparative low-bitwidth integers to ensure that back-propagation requires integer numbers, which reduces the hardware requirement of training. \citet{park2019distilling} showed one way to generate a high-quality machine learning model based on on-device model parameters, output, and data aggregation, while \citet{jiang2019model} proposed a method for efficient learning using pruned models. Another exciting approach presented in \cite{leroy2019federated} discussed training higher capacity models with fewer parameters. Particularly, energy-efficient learning on resource-constrained devices is essential if the size and number of features of the training dataset are large. In such a case, generating a higher-quality model by considering fewer parameters and performing on-device training could be a challenge.

\subsection{Scheduling}
In synchronous FL, all clients interact with server at the same time, while in asynchronous FL, the training period can be different. Hence, it is essential to determine the training period for all local participants, which we call \textbf{scheduling}. If we consider resource-constrained clients, then it is not an ideal solution to carry out scheduling frequently. Rather, an optimized scheduling period would cost minimal energy consumption and less bandwidth. To achieve this, parallel training sessions can be avoided due to high resource consumption, and a worker queue can be maintained for on-device multi-tenant systems. Moreover, clients should not perform scheduling tasks when they possess old data. It may be possible that older data are repeatedly used for training, while newer data are omitted \cite{bonawitz2019towards}. Old data will not give us much variation to the model parameter, and resources will be wasted without model improvement. In addition, any client may frequently use a particular app that provides malicious data, and identification of such app usage is also a challenge. Thus, it is necessary to execute scheduling after filtering out which data should be used for training.

\subsection{Stragglers}
In FL, one main reason for performance bottleneck is the presence of straggler clients. While each client is responsible for generating a model with their data and sharing that with the server after a certain period, a straggler client may fail to share its model with the server at a proper time convenience. Due to this, the central server needs to wait until all the straggler clients share their model. Hence, the overall training procedure of the clients is delayed. One solution to avoid straggler clients is to select competent clients based on their resource-availability. \citet{das2019privacy} proposed a solution to acknowledge the computational power, i.e., overall resource utilization of the clients after each local update. By observing each client's resource utilization, a predictive model can be formed for adjusting the local computation of the clients. Another option is to use asynchronous training, which, however, is challenging in the presence of non-independent/identically distributed (non-IID) data among clients (see Section~\ref{subsec:asynchronous}).

\subsection{Labelling Unlabelled Data}
In the FL system, most existing techniques considered that data are labeled. However, data collected by local devices can be unlabeled due to connection or communication issues and can be mislabeled. \citet{gu2019reaching} designed a framework to identify mislabeled data, while \citet{lim2019federated} proposed a solution to put a label on  unlabeled data using collaborative learning with neighboring devices. However, it is challenging to label data in real-time for resource-constrained devices, as a client may have power limitations or other issues. 

\section{Open Issues of Federated Learning Algorithms and Hardware Developments}
\label{sec:5}
\begin{figure}[htb!]
\setlength{\belowcaptionskip}{-20pt}
\begin{center}
  \includegraphics[width=0.65\linewidth]{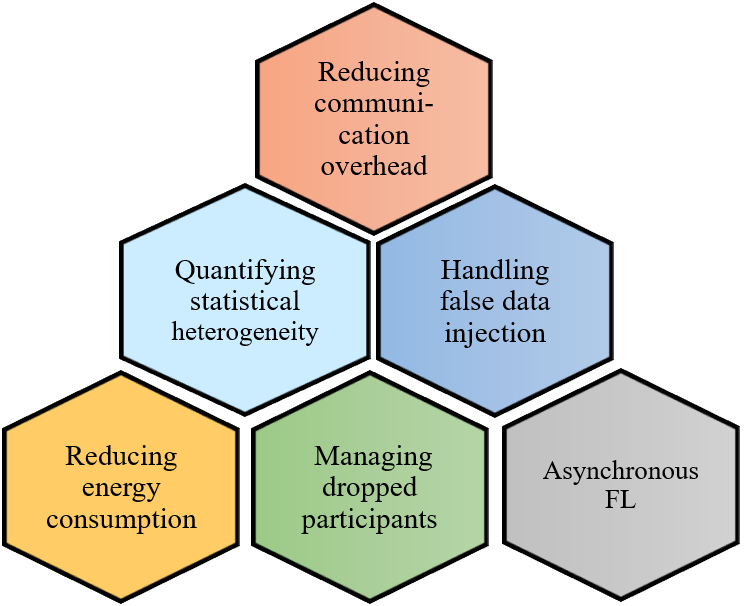}
  \vspace{-0.1in}
    \caption{Open issues and future directions of the federating learning theory and applications.}
    \label{fig:3}  
    \end{center}
\end{figure}

 In the previous section, we discussed the core challenges for resource-constrained IoT clients while deploying FL.  There still exist some aspects that need to be addressed (see Figure~\ref{fig:3}) and can be considered as open research problems. In this section, we highlight promising future trends.

\subsection{Deploying Existing Algorithm to Reduce Communication Overhead}
Existing methods in the literature that are proposed to reduce communication overhead can be categorized into decentralized training, compression, and local updating. Integration of these methods can generate an optimized FL platform with fast convergence time. For instance, infusing redundancy amongst the client dataset was proposed in \cite{haddadpour2019trading} for bringing diversity and reaching convergence in a shorter time.  \citet{chen2019joint} presented a joint learning framework by quantifying the impact of wireless factors on clients in FL environment. Still, further researches are needed to attain minimal communication overhead at scale.

\subsection{Guarantee Convergence in Asynchronous Federated Learning}
\label{subsec:asynchronous}
Most existing studies and implementations consider synchronous FL, where the progress of the iteration round's training period depends on the slowest device within the network. This means synchronous FL has a direct effect on the overall performance of the system, although it guarantees convergence. In asynchronous FL, the participant can join the training round even in the middle of the training progress. It also ensures scalability although it does not guarantee convergence \cite{sprague2018asynchronous}. Considering the effectiveness of asynchronous FL is one of the core research issues to formulate a method for ensuring convergence during the asynchronous training of clients.

\subsection{Quantification of Statistical Heterogeneity}
In an IoT environment, the data collected by local devices are inherently non-IID; thus, they may have a discrepancy in terms of the number of samples and dataset structure. It is challenging to quantify the level of heterogeneity in the system before the training begins. A local dissimilarity based strategy was designed in \cite{eliazar2010measuring} to quantify heterogeneity, but it cannot quantify heterogeneity before training starts. If we have a mechanism of quantifying heterogeneity at initialization, the system can be configured accordingly to allow more efficient training. 

\subsection{Handling False Data Injection}
In a distributed system, the local devices are responsible for generating their model using the raw data they extracted. But, somehow, if the client constructs its model using false data that are injected during data extraction, then the generated model will cause an erroneous update to the global model. This opens up a new research direction to identify false data efficiently, particularly using resource-constrained devices with the limitations discussed earlier.

{\color{black}\subsection{On-device Training}
As discussed previously, devices in the IoT domain can have extremely limited capabilities. As a result, it becomes extremely important to do training and inference on the device and, thus, limit the interaction with servers via energy-inefficient communication methods. However, training on a device leads to two problems. First, finding a model with enough capacity that can run on devices having small size of memory, while still capturing the complexity of the data can be hard. \citet{kumar2017resource} and \citet{thakker2019compressing} 
solved these problems for inference, but did not discuss training the models on the device. Second, training can require a much larger computational and memory capacity than these simple clients can provide. Section~\ref{subsec:energyEfficientTraining} discussed some techniques which either work on specialized neuromorphic or FPGA hardware or do not meet the aggressive constraints found in the IoT domain. Solving this dual problem is paramount. \citet{thakker2019compressing} provided a potential direction in this regard. They suggested a new network architecture for IoT application to create high capacity models with 15-38x fewer parameters than a traditional model used for those applications. Exploring such new architectures and enabling them to learn over the severely resource-constrained FL environment is an unexplored domain.}

\subsection{Managing Dropped Participants}
In the resource-constrained FL system, clients might have heterogeneous resources, i.e. variable bandwidth, limited battery life and variant transmission power. As a result, any client within a network can be disconnected during the communication with the server. Recent studies widely assumed that all the participants are available and connected with the server throughout the process. But, in real-life, any client can go offline due to the non-availability of resources. Eventually, the disconnection of a significant number of clients can degrade the convergence speed and model accuracy. 

\vspace{-0.05in}
\section{Future Directions}
{{\color{black}

FL is a recently invented technique and an active ongoing research area. After analyzing the potential challenges of implementing FL in resource-constrained clients in Section \ref{sec:4} and discussing some open issues along with potential solutions in Section \ref{sec:5}, we figure-out some future directions that need to be highlighted. In this section, we point out those future directions. 
}} 
{\color{black}

  {\color{black}

  \noindent $\bullet$ In the FL environment, some clients may generate more data (i.e., heavier use of an application by a particular user) than other clients within the underlying network of decision-making entities. This may lead to varying amounts of local data during the training period, and we can not represent any client dataset as population distribution. Handing such discrepancy in \textbf{local training dataset} requires further research.}

  
  \noindent $\bullet$ To ensure convergence in a non-IID scenario, particularly for asynchronous learning, loss functions of the \textbf{non-convex problem} need to be considered, and supportive algorithms should be proposed. 
  
  
  \noindent $\bullet$ In the FL scenario,  selection of suitable cluster heads and maintaining coordination within the overall system need more investigation to ensure \textbf{energy efficiency}. {\color{black}It may be possible that a cluster of clients agrees to generate an aggregated model using their local data and disseminate it to the central server.  The aggregated model can be passed by the leader client, and hence, it can reduce energy consumption.   In order to reduce energy consumption, we can consider a group of clients as a cluster in a distributed structure,  and select a proficient client that will act as a leader client. That client will be responsible for interaction with the central server in an asynchronous fashion, while the interaction inside each cluster's agents is conducted in a synchronous fashion. This reduces the energy consumption of the overall system by avoiding unnecessary communication among all agents and the central server. 
} 
  
  \noindent $\bullet$ A \textbf{device-centric wake-up mechanism} can be built through which the clients can automatically understand the period to interact with the server. This functionality will help resource-constrained clients, particularly in asynchronous FL, where a client needs to wait for other neighbor clients to send their models to the server. By building such a mechanism, the energy consumption rate of the clients can be reduced by a significant margin.
  

  \noindent $\bullet$ The resource-constrained IoT clients may need to perform more interaction with the server due to \textbf{statistical heterogeneity} in the system. So, it is necessary to evaluate an efficient method for identifying the statistical heterogeneity even before the training phase and to avoid idiosyncratic situations due to data sample variation.
  
  \noindent $\bullet$ In terms of {scalability}, frequent drop-out of the participant is a significant bottleneck. A new approach should be adapted to make the FL system more \textbf{robust to frequent drop-outs}. One solution could be predicting or identifying the probability of a participant disconnection. Further, a dedicated connection (e.g., cellular connection) can be provided as an incentive to avoid connection drop-out \cite{lim2019federated}.  Moreover, a structure can be maintained while designing a protocol so that drop-out participants can try to make a connection multiple times during the long-running model-aggregation. The issue regarding intermittent client availability at scale is not addressed in prior works.


  \noindent $\bullet$ Due to \textbf{mobility} of clients, new clients may join  a network that is more competent than the existing clients, and any client may leave the network during communication that can hamper the model training. {\color{black}Besides, because of mobility, there may be a large number of clients in some areas while other areas may not have enough clients to generate a feasible model.} Handling such situations by considering both mobility and bandwidth ability of clients can be a research direction.

  \noindent $\bullet$ The optimal communication degree in the FL implementation is still ongoing research, and there is not yet a  deterministic algorithm to identify the \textbf{globally-optimum communication topology}. Although divide-and-conquer and one-shot communication methods are discussed in \cite{zhang2015divide}, these schemes are not well-suited for heterogeneous resource-constrained clients. This specifically leads to challenges related to FL implementation in heterogeneous IoT systems with time-varying communication topology, such as distributed mobile sensor networks.  Although one-shot and few-shot FL approaches are proposed in \cite{guha2019one},  extensive practical evaluation is needed to identify a solution to the optimal communication topology problem.   In the context of wireless networks, the distribution of \textbf{fair resources} has been studied extensively. While optimizing the utilities may give us higher throughput, unfair resource allocations may cause inadequate service facilities. The global model can be considered as a resource for providing service to clients. If we use asynchronous FL, then any client may receive a pre-assigned \textbf{fairness} to modify its objective function during the training period. We can handle trade-offs between fairness and other metrics (e.g., average accuracy) by tuning the parameter. Still, more theoretical and practical analysis needs to be conducted to optimize resource distribution, particularly for resource-constrained devices.

 \noindent $\bullet$ A \textbf{blockchain} paradigm can be constructed to make the FL clients' communication more robust and secure. The model update and exchange of resource-constrained IoT clients can be verified by blockchain. \citet{kim2019blockchained} and \citet{xu2020ndss} proposed a blockchain-based on-device FL, but they did not consider resource-constrained clients. How the resource-constrained IoT clients perform block traversing, select miners as well as leader-client, ensure atomicity, and reach a consensus for FL scenario is a future direction. 
 
 {\color{black}\noindent $\bullet$ Designing an \textbf{incentive mechanism for transparent participation} is required in FL. As participants may be resource-bounded or business competitors, it is essential to develop an approach that effectively divides the earnings in order to enhance the long-term participation of the clients. Furthermore, how to defend against adversarial client data owners and optimize the non-adversarial owner participation for ensuring security needs to be explored. }
 
 {\color{black}\noindent $\bullet$ The solitary computational model of FL can lead us to build a more refined trust-based model. As it is challenging in terms of \textbf{security} to select a participant for the training phase, a \textbf{trust-based model} can reduce extra communication overhead. Typically, we assume that the server is operated by a non-adversarial entity. This server can analyze the behavior of participants and leverage a trust model. According to the trust score, an incentive mechanism can be designed, and that trust model can assist us when the number of participants is significant. This opens up a new research direction to explore.}
 
}

\section{Conclusion}
In this paper, we conducted a comprehensive survey on FL, particularly for resource-constrained IoT devices, and highlighted the ongoing research related to this area. We began by discussing the importance of leveraging FL for resource-constrained IoT clients and discussed some previous works that considered resource scarcity during the implementation of FL system. We highlighted the background, including the working procedure of FL, and explored existing FL applications by discussing their importance in conducting model training locally. Further, we focused on core challenges of implementing FL, particularly for resource-constrained devices by considering hardware limitation, communication expense, client behavior, statistical data variation and labeling, and energy-efficient training. Finally, we emphasized the need for future directions for contriving new FL algorithms in terms of currently open issues and designing the latest hardware considering resource-constrained challenges, especially in an FL scenario.

\clearpage
\bibliographystyle{named}
\bibliography{ijcai20}

\begin{thebibliography}{}

\bibitem[\protect\citeauthoryear{Bonawitz \bgroup \em et al.\egroup
  }{2019}]{bonawitz2019towards}
Keith Bonawitz, Eichner, et~al.
\newblock Towards federated learning at scale: System design.
\newblock {\em arXiv preprint arXiv:1902.01046}, 2019.

\bibitem[\protect\citeauthoryear{Chen \bgroup \em et al.\egroup
  }{2018}]{chen2018federated}
Fei Chen, Zhenhua Dong, et~al.
\newblock Federated meta-learning for recommendation.
\newblock {\em arXiv:1802.07876}, 2018.

\bibitem[\protect\citeauthoryear{Chen \bgroup \em et al.\egroup
  }{2019}]{chen2019joint}
Mingzhe Chen, Zhaohui Yang, et~al.
\newblock A joint learning and communications framework for federated learning
  over wireless networks.
\newblock {\em arXiv:1909.07972}, 2019.

\bibitem[\protect\citeauthoryear{Cui \bgroup \em et al.\egroup
  }{2018}]{cui2018survey}
Laizhong Cui, Shu Yang, et~al.
\newblock A survey on application of machine learning for internet of things.
\newblock {\em J. M. L. Cybernetics}, 9(8):1399--1417, 2018.

\bibitem[\protect\citeauthoryear{Das and Brunschwiler}{2019}]{das2019privacy}
Anirban Das and Thomas Brunschwiler.
\newblock Privacy is what we care about: Experimental investigation of
  federated learning on edge devices.
\newblock In {\em AIChallengeIoT}, 2019.

\bibitem[\protect\citeauthoryear{Eliazar and
  Sokolov}{2010}]{eliazar2010measuring}
Iddo~I Eliazar and Igor~M Sokolov.
\newblock Measuring statistical heterogeneity: The pietra index.
\newblock {\em Physica A: Stat. Mech. App.}, 389(1):117--125, 2010.

\bibitem[\protect\citeauthoryear{Gu \bgroup \em et al.\egroup
  }{2019}]{gu2019reaching}
Zhongshu Gu, Hani Jamjoom, et~al.
\newblock Reaching data confidentiality and model accountability on the
  caltrain.
\newblock In {\em IEEE DSN}, 2019.

\bibitem[\protect\citeauthoryear{Guha \bgroup \em et al.\egroup
  }{2019}]{guha2019one}
Neel Guha, Ameet Talwlkar, et~al.
\newblock One-shot federated learning.
\newblock {\em arXiv:1902.11175}, 2019.

\bibitem[\protect\citeauthoryear{Haddadpour \bgroup \em et al.\egroup
  }{2019}]{haddadpour2019trading}
Farzin Haddadpour, Mohammad~Mahdi Kamani, et~al.
\newblock Trading redundancy for communication: Speeding up distributed sgd for
  non-convex optimization.
\newblock In {\em ICML}, 2019.

\bibitem[\protect\citeauthoryear{Hard \bgroup \em et al.\egroup
  }{2018}]{hard2018federated}
Andrew Hard, Kanishka Rao, Rajiv Mathews, Ramaswamy, et~al.
\newblock Federated learning for mobile keyboard prediction.
\newblock {\em arXiv:1811.03604}, 2018.

\bibitem[\protect\citeauthoryear{Huang and others}{2013}]{huang2013depth}
Junxian Huang et~al.
\newblock An in-depth study of lte: effect of network protocol and application
  behavior on performance.
\newblock {\em ACM SIGCOMM CCR}, 43(4):363--374, 2013.

\bibitem[\protect\citeauthoryear{Huang \bgroup \em et al.\egroup
  }{2018}]{huang2018loadaboost}
Li~Huang, Yifeng Yin, et~al.
\newblock Loadaboost: Loss-based adaboost federated machine learning on medical
  data.
\newblock {\em arXiv:1811.12629}, 2018.

\bibitem[\protect\citeauthoryear{Jiang \bgroup \em et al.\egroup
  }{2019}]{jiang2019model}
Yuang Jiang, Shiqiang Wang, et~al.
\newblock Model pruning enables efficient federated learning on edge devices.
\newblock {\em arXiv:1909.12326}, 2019.

\bibitem[\protect\citeauthoryear{Kim \bgroup \em et al.\egroup
  }{2019}]{kim2019blockchained}
Hyesung Kim, Jihong Park, et~al.
\newblock Blockchained on-device federated learning.
\newblock {\em IEEE Communications Letters}, 2019.

\bibitem[\protect\citeauthoryear{Kone{\v{c}}n{\`y} \bgroup \em et al.\egroup
  }{2016}]{konevcny2016federated}
Jakub Kone{\v{c}}n{\`y}, H~Brendan McMahan, et~al.
\newblock Federated learning: Strategies for improving communication
  efficiency.
\newblock {\em arXiv:1610.05492}, 2016.

\bibitem[\protect\citeauthoryear{Kumar \bgroup \em et al.\egroup
  }{2017}]{kumar2017resource}
Ashish Kumar, Saurabh Goyal, et~al.
\newblock Resource-efficient machine learning in 2 kb ram for the internet of
  things.
\newblock In {\em ICML}, 2017.

\bibitem[\protect\citeauthoryear{Leroy and others}{2019}]{leroy2019federated}
David Leroy et~al.
\newblock Federated learning for keyword spotting.
\newblock In {\em IEEE ICASSP}, 2019.

\bibitem[\protect\citeauthoryear{Li \bgroup \em et al.\egroup
  }{2018}]{li2018learning}
He~Li, Kaoru Ota, et~al.
\newblock Learning iot in edge: Deep learning for the internet of things with
  edge computing.
\newblock {\em IEEE network}, 32(1):96--101, 2018.

\bibitem[\protect\citeauthoryear{Li \bgroup \em et al.\egroup
  }{2019}]{li2019federated}
Tian Li, Anit~Kumar Sahu, et~al.
\newblock Federated learning: Challenges, methods, and future directions.
\newblock {\em arXiv:1908.07873}, 2019.

\bibitem[\protect\citeauthoryear{Lim \bgroup \em et al.\egroup
  }{2019}]{lim2019federated}
Wei Yang~Bryan Lim, Nguyen~Cong Luong, et~al.
\newblock Federated learning in mobile edge networks: A comprehensive survey.
\newblock {\em arXiv:1909.11875}, 2019.

\bibitem[\protect\citeauthoryear{Ma \bgroup \em et al.\egroup
  }{2017}]{ma2017distributed}
Chenxin Ma, Jakub Kone{\v{c}}n{\`y}, et~al.
\newblock Distributed optimization with arbitrary local solvers.
\newblock {\em Optimization Methods and Software}, 32(4):813--848, 2017.

\bibitem[\protect\citeauthoryear{McMahan \bgroup \em et al.\egroup
  }{2016}]{mcmahan2016communication}
H~Brendan McMahan, Eider Moore, et~al.
\newblock Communication-efficient learning of deep networks from decentralized
  data.
\newblock {\em arXiv:1602.05629}, 2016.

\bibitem[\protect\citeauthoryear{Park \bgroup \em et al.\egroup
  }{2019}]{park2019distilling}
Jihong Park, Shiqiang Wang, et~al.
\newblock Distilling on-device intelligence at the network edge.
\newblock {\em arXiv:1908.05895}, 2019.

\bibitem[\protect\citeauthoryear{Pete Warden}{2018}]{pete}
What does it take to train deep learning models on-device?, 2018.

\bibitem[\protect\citeauthoryear{Sprague \bgroup \em et al.\egroup
  }{2018}]{sprague2018asynchronous}
Michael~R Sprague, Amir Jalalirad, et~al.
\newblock Asynchronous federated learning for geospatial applications.
\newblock In {\em ECML-PKDD}, 2018.

\bibitem[\protect\citeauthoryear{Thakker \bgroup \em et al.\egroup
  }{2019}]{thakker2019compressing}
Urmish Thakker, Jesse Beu, et~al.
\newblock Compressing rnns for iot devices by 15-38x using kronecker products.
\newblock {\em arXiv preprint arXiv:1906.02876}, 2019.

\bibitem[\protect\citeauthoryear{{Wang} \bgroup \em et al.\egroup
  }{2019}]{wang2018edge}
Shiqiang {Wang}, Tiffany {Tuor}, et~al.
\newblock Adaptive federated learning in resource constrained edge computing
  systems.
\newblock {\em IEEE JSAC}, 37(6):1205--1221, 2019.

\bibitem[\protect\citeauthoryear{Wu \bgroup \em et al.\egroup
  }{2018}]{wu2018training}
Shuang Wu, Guoqi Li, et~al.
\newblock Training and inference with integers in deep neural networks.
\newblock {\em arXiv:1802.04680}, 2018.

\bibitem[\protect\citeauthoryear{Xu \bgroup \em et al.\egroup
  }{2019a}]{xu2019exploring}
Zichen Xu, Li~Li, et~al.
\newblock Exploring federated learning on battery-powered devices.
\newblock In {\em ACM TURC}, 2019.

\bibitem[\protect\citeauthoryear{Xu \bgroup \em et al.\egroup
  }{2019b}]{xu2019elfish}
Zirui Xu, Zhao Yang, et~al.
\newblock Elfish: Resource-aware federated learning on heterogeneous edge
  devices.
\newblock {\em arXiv:1912.01684}, 2019.

\bibitem[\protect\citeauthoryear{Xu \bgroup \em et al.\egroup
  }{2020}]{xu2020ndss}
Ronghua Xu, Yu~Chen, and Jian Li.
\newblock Micro{FL}: A lightweight, secure-by-design edge network fabric for
  decentralized {IoT} systems.
\newblock In {\em NDSS}, 2020.

\bibitem[\protect\citeauthoryear{Yang \bgroup \em et al.\egroup
  }{2018}]{yang2018applied}
Timothy Yang, Galen Andrew, et~al.
\newblock Applied federated learning: Improving google keyboard query
  suggestions.
\newblock {\em arXiv:1812.02903}, 2018.

\bibitem[\protect\citeauthoryear{Yang \bgroup \em et al.\egroup
  }{2019}]{yang2019federated}
Qiang Yang, Yang Liu, et~al.
\newblock Federated machine learning: Concept and applications.
\newblock {\em ACM Trans. on TIST}, 10(2):12, 2019.

\bibitem[\protect\citeauthoryear{Zhang \bgroup \em et al.\egroup
  }{2015}]{zhang2015divide}
Yuchen Zhang, John Duchi, et~al.
\newblock Divide and conquer kernel ridge regression: A distributed algorithm
  with minimax optimal rates.
\newblock {\em JMLR}, 16(1):3299--3340, 2015.

\end{thebibliography}

\end{document}